\documentclass[conference]{IEEEtran}
\IEEEoverridecommandlockouts
\usepackage{cite}
\usepackage{tabularx,booktabs}
\usepackage{amsmath,amssymb,amsfonts}
\usepackage{algorithmic}
\usepackage{graphicx}
\usepackage{textcomp}
\usepackage{xcolor}
\def\BibTeX{{\rm B\kern-.05em{\sc i\kern-.025em b}\kern-.08em
    T\kern-.1667em\lower.7ex\hbox{E}\kern-.125emX}}
\begin{document}

\title{Graph Neural Ordinary Differential Equations-based method for Collaborative Filtering\\

}

\author{\IEEEauthorblockN{Ke Xu}
\IEEEauthorblockA{\textit{University of Illinois Chicago} \\
Chicago, USA \\
kxu25@uic.edu}
\and
\IEEEauthorblockN{Yuanjie Zhu}
\IEEEauthorblockA{\textit{University of Illinois Chicago} \\
Chicago, USA \\
yzhu224@uic.edu}
\and
\IEEEauthorblockN{Weizhi Zhang}
\IEEEauthorblockA{\textit{University of Illinois Chicago} \\
Chicago, USA \\
wzhan42@uic.edu}
\and
\IEEEauthorblockN{Philip S. Yu}
\IEEEauthorblockA{\textit{University of Illinois Chicago} \\
Chicago, USA \\
psyu@uic.edu}
}

\maketitle

\begin{abstract}
Graph Convolution Networks (GCNs) are widely considered state-of-the-art for collaborative filtering. Although several GCN-based methods have been proposed and achieved state-of-the-art performance in various tasks, they can be computationally expensive and time-consuming to train if too many layers are created.
However, since the linear GCN model can be interpreted as a differential equation, it is possible to transfer it to an ODE problem. This inspired us to address the computational limitations of GCN-based models by designing a simple and efficient NODE-based model that can skip some GCN layers to reach the final state, thus avoiding the need to create many layers. In this work, we propose a Graph Neural Ordinary Differential Equation-based method for Collaborative Filtering (GODE-CF). This method estimates the final embedding by utilizing the information captured by one or two GCN layers. To validate our approach, we conducted experiments on multiple datasets. The results demonstrate that our model outperforms competitive baselines, including GCN-based models and other state-of-the-art CF methods. Notably, our proposed GODE-CF model has several advantages over traditional GCN-based models. It is simple, efficient, and has a fast training time, making it a practical choice for real-world situations.
\end{abstract}

\begin{IEEEkeywords}
recommendation, neural ordinary differential equations, graph neural ordinary differential equations, collaborative filtering, graph convolutional networks
\end{IEEEkeywords}

\section{Introduction}
In recent years, machine learning tasks on graph-structured data have become increasingly popular, and Graph Convolutional Networks (GCNs) \cite{kipf_semi-supervised_2017, zhou_graph_2020, zhang_graph_2019, berg_graph_2017} have emerged as a powerful tool for these tasks. GCNs are effective in various tasks, particularly in recommendation systems, due to their simplicity and effectiveness.
The core idea of GNNs is to update each node embedding iteratively using multiple graph propagation layers. This process aggregates node embeddings from both their neighbors and themselves. Essentially, GNNs learn the embeddings of nodes by leveraging the graph's structure. In practice, it has been observed that most tasks require only two or three layers \cite{qu_gmnn:_2020} to achieve good results.
Using a larger number of layers in GCNs may result in inferior performance, as the model becomes increasingly complex and its parameters more difficult to optimize \cite{kipf_semi-supervised_2017, zhou_graph_2020, li_deeper_2018}.

Collaborative Filtering(CF)\cite{chen_revisiting_2020,chen_survey_2018,wang_neural_2019,rendle_bpr:_2009,he_neural_2017, yang_survey_2014, ebesu_collaborative_2018, wu_collaborative_2016} is a popular approach used in recommender systems to predict users' interests based on their past behavior or preferences. The goal is to learn user and product embeddings and compute dot-products for recommendations. However, this approach has long been a research problem in the field of recommender systems.
And Graph Convolutional Networks (GCNs) can effectively capture the relationships between users and items in a graph, enhancing the performance of CF. This approach represents users and products as nodes in a graph, with edges between them representing interactions or relationships. A GCN will learn node embeddings by iteratively aggregating information from neighboring nodes in the graph. GCNs have shown promising results in improving the performance of traditional CF methods and are increasingly used in various recommendation systems. It is particularly useful in scenarios where user-item interactions are complex and sparse, as they can leverage graph structure to capture higher-order relationships between nodes.

Several studies\cite{he_lightgcn:_2020, wu_simplifying_2019} have found that linear GCN architectures with layer combination outperform non-linear GCN \cite{he_neural_2017} architectures for collaborative filtering (CF). Additionally, a linear GCN can be easily interpreted as differential equations, this concept\cite{Chen_node, deng_continuous_2019} has led to a Neural Ordinary Differential Equations(NODEs)-based CF method called LT-OCF\cite{choi_lt-ocf:_2021}. This method demonstrates the suitability of NODEs-based approaches for CF. The main idea behind LT-OCF is to develop a purely continuous version of the GCN layer. The architecture can be viewed as a continuous version of LightGCN\cite{he_lightgcn:_2020} with a learnable number of layers.

Recent study\cite{Chen_node} on neural ordinary differential equations(NODEs) suggests that discrete layers with residual connections may not be the only perspective for building neural networks. Instead, this line of research proposes that discrete layers with residual connections can be understood as a discretization of a continuous ordinary differential equation(ODE). This approach is more memory-efficient since NODEs can significantly reduce the required number of parameters when building neural networks and models\cite{pinckaers_neural_2019}, and can also model the dynamics of hidden states more smoothly. The implications of this approach are significant because it opens up new possibilities for constructing neural networks that are more flexible and efficient. 

The concept behind neural ordinary differential equations(NODEs) are to learn implicit differential equations from data. NODEs calculate the output $h(t_1) = h(t_0)+\int^{t_1}_{t_0}f(h(t), \theta) dt$, where $f(h(t), \theta)$ is a neural network. $\theta$ is trained from data. And $t_0$ is the staring time, $t_1$ is ending time. The resulting $h(t_1)$ values provide an approximation of the solution to the differential equation. There are various ODE solvers that can solve integral problems, and they can also generalize different types of neural network architectures\cite{Chen_node}. For instance, the explicit Euler method for soloving ODE problems can be viewed as a residual connection\cite{Chen_node,choi_lt-ocf:_2021}. Meanwhile, the fourth-order Runge-Kutta(RK4) ODE solver can be seen as a dense convolutional network\cite{zhu_convolutional_2023}.

Because of the conceptual connections between Graph Neural Networks (GNNs)\cite{keramatfar_graph_2022,kipf_semi-supervised_2017} and NODEs, which have led to the development of a continuous version of GNNs known as Continuous Graph Neural Networks (CGNNs)\cite{xhonneux_continuous_2020}. CGNNs extend existing discrete GNNs to the continuous case by defining the evolution of node embeddings using ordinary differential equations (ODEs). A CGNN calculates final node embeddings $h(t_1) = h(0) + \int_{0}^{t_1} (A-I)h(t) dt$, where $A$ is the adjacency matrix. The node embeddings of $h(t_1)$ at step $t_1$ can be viewed as all the information propagated up to $t_1$ steps with the node embeddings $h(t)$. In this way $t_1$ can be viewed as the number of layers for GNNs. The method we mentioned earlier, LT-OCF, utilizes a similar idea to CGNNs.

Different than CGNNs, Graph Neural Ordinary Differential Equations(GODEs)\cite{poli_graph_2021} do not develop a continuous message-passing layer. Instead, they use the NODE framework and parameterize the derivative function directly using multiple GCN layers. Since GCNs perform message passing to derive final states, it is similar to solving an initial value problem for an ODE. The final states of node embeddings $h(t_1)$ can be calculated using $h(t_1) = h(t_0) + \int^{t_1}_{t_0} G(h(t)) dt$, where $G(h(t))$ is a multiple GCN layer with the node embeddings at state $t$, $h(t)$. GODEs retains the concept of the number of layers $n$ and views the time step $t_1$ as another hyperparameter. To put it simply, GODEs seek to estimate the end node embeddings by utilizing data captured by just 1 or 2 GCN layers.

Inspired by the concept of Graph Neural Ordinary Differential Equations (GODEs), we propose a new method called Graph Neural Ordinary Differential Equations-based Collaborative Filtering (GODE-CF). GODE-CF is designed to estimate the final embeddings by utilizing an ODE. Unlike many other GCN methods for Collaborative Filtering (CF), GODE-CF does not utilize layer combinations. Instead, we will use the output of the ODE as the final embeddings. Our proposed method has several advantages over other linear GCN models. It has a faster training time without sacrificing accuracy, and it achieves state-of-the-art accuracy in CF tasks.

\begin{figure*}[htbp]
\centerline{\includegraphics[scale=0.55]{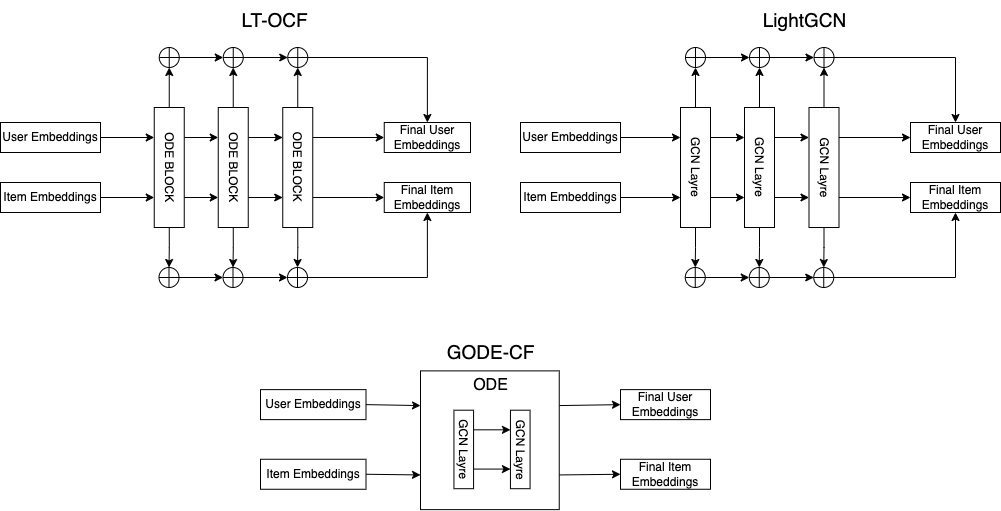}}
\caption{The architectures of LT-OCF, LightGCN and GODE-CF. \textbf{Left} figure: a LT-OCF with three ODE blocks; \textbf{Right} figure: a LightGCN with three layers, and $\bigoplus$ is the layer combination; \textbf{Middle} figure: GODE-CF with two layers.}
\label{fig:arch}
\end{figure*}

Figure \ref{fig:arch} shows the architecture of GODE-CF and a comparison between LightGCN and LT-OCF. To evaluate the performance of GODE-CF, we use four public review datasets and compare them with state-of-the-art methods, such as LT-OCF\cite{choi_lt-ocf:_2021}, LightGCN\cite{he_lightgcn:_2020} and UltraGCN\cite{mao_ultragcn:_2021}. The results show that our method consistently outperforms these methods in all datasets. Furthermore, we demonstrate the effectiveness of our method by showing that it can be trained faster than most GCN-based methods. Additionally, we investigate the impact of the choice of solver for ODEs on different datasets. Our contributions can be summarized as follows:
\begin{itemize}
\item We propose a novel architecture for CF, Graph Neural Ordinary Differential Equations-based Collaborative Filtering (GODE-CF), which incorporates both GCN and GODE components..

\item We demonstrate that our method outperforms state-of-the-art methods in all datasets.

\item We show that our method can be trained faster than most GCN-based methods.

\item We investigate the impact of the choice of solver for ODEs on different datasets.
\end{itemize}

\section{PRELIMINARIES}
\subsection{Graph Convolution Networks(GCN)}
Graph convolution network(GCN) is variant of Graph Neural Networks(GNNs). The basic idea of GCNs is to learn embeddings for nodes by smoothing features over the graph\cite{kipf_semi-supervised_2017, zhou_graph_2020, zhang_graph_2019, berg_graph_2017}. For each layer, it aggregates the features of neighbors as the new embeddings of the target node. The propagation rule can be written as:
\begin{equation}
    H_{n+1} = \sigma(AGG(W, A, H_n))
\end{equation}
Where $AGG$ is aggregation function, $H_n$ is node embedding for the $n$th layer, $H_{n+1}$ is node embedding for $n+1$th layer, $W$ is the transformation matrix, and $A$ is adjacency matrix of the graph. 
However, there are studies\cite{kipf_semi-supervised_2017} have shown that the GCN can only learn graph moments of a specific power of the Graph. To overcome this limitation, many GCN models\cite{kipf_semi-supervised_2017,he_neural_2017,he_lightgcn:_2020} concatenate the node embeddings of each layer, which allows the model to capture information from different powers of $A$ and improves its ability to model complex graph structures. 

\subsection{Neural Ordinary Differential Equations(NODEs)}
Neural Ordinary Differential Equations(NODEs)\cite{Chen_node} is a method that aims to model continuous dynamics on hidden state of neural networks. This is achieved by characterizing dynamics through an ordinary differential equation (ODE), which is parameterized by a neural network. The goal of this approach is to learn implicit differential equations from data. One of the key advantages of Neural ODEs is that they provide a flexible framework for modeling complex systems. By using ODEs, it is possible to model systems that exhibit continuous behavior. Additionally, by using neural networks to parameterize the ODEs, it is possible to capture complex patterns in the data that may be difficult to model using discrete methods. Neural ordinary differential equations calculate $h(t_1) = h(t_0) + \int_{t_0}^{t_1} f(h(t), \theta )dt$, where $f$ is a neural network parameterized by $\theta$ that approximates $\frac{dh(t)}{dt}$, to derive $h(t_1)$ from $h(t_0)$. $\theta$ is trained from data. The variable $t_0$ and $t_1$ are staring time and ending time, commonly set $t_0=0$ and then $t_1$ can be viewed as the number of layers in a neural networks, whereas it can be any non-negative real number in NODEs. In order to solve ODE problem, there are different ODE solvers that can be used to solve the ODE problem, such as Euler, RK2, RK4, and they can also be viewed as various neural network architectures. For example, the general form of the residual connection can be written as $h(t+1)=h(t) + f(h(t);\theta)$, which is same as the explicit Euler method for solving ODE problems. In general, the choice of ODE solver depends on the specific problem being solved. For instance, the explicit Euler method is simple and easy to implement, but it can be unstable. On the other hand, the fourth-order Runge-Kutta(RK4) method is more stable and accurate, but it can be computationally expensive\cite{poli_graph_2021, choi_lt-ocf:_2021}.

\section{RELATED WORK}
\subsection{Continuous Graph Neural Networks(CGNNs)}
Deep learning has traditionally been dominated by discrete models. However, there are some studies\cite{haber_stable_2017, Chen_node} have proposed treating neural networks as equipped with a continuum of layers, rather than a discrete set of layers. This approach enables a reformulation of the forward and backward pass as the solution to the initial value problem of an ODE. Such approaches allow direct modeling of ODEs and can guide discovery of novel general purpose deep learning models. This concept also build some connections between GCNs and traditional dynamical systems. Continuous Graph Neural Networks (CGNNs)\cite{xhonneux_continuous_2020} are an architecture that models continuous dynamics on node embeddings. First, it employ a neural encoder to project each node into a latent space based on its features. Afterwards, the encoded nodes are treated as the initial value $H(0)$, and an ODE is designed to define the continuous dynamics on node embeddings. This allows CGCNs to capture complex patterns of interactions between nodes over time. Finally, the node embeddings obtained at the ending time $t_1$ can be used for downstream applications such as node classification. Motivated by diffusion-based methods on graphs, an effective way for characterizing the propagation process is to use the following propagation equations:
\begin{equation}
    H_{n+1} = AH_n +H_0
\end{equation}
where $H_n$ is the embeddings for all the nodes at stage $n$, and $H_0$ is the initial embeddings. Intuitively, each node at stage $n + 1$ learns the node information from its neighbors through $AH_n$ and remembers its original node features through $H_0$. This way is to learn the graph structure without forgetting the original node features.
And then, the formula can be derived as follows:
\begin{equation}
    H_n = (\sum_{i=0}^n A^i)H_0 = (A-I)^{-1}(A^{n}-I)H_0
\end{equation}
where the embeddings $H_n$ at stage $n$ incorporates all the information propagated up to $n$ steps with the initial embeddings $H_0$. With the idea of NODEs, we can replace the discrete step $n$ with a continuous variable $t$, and use an ODE to characterize such a continuous propagation dynamic. This allows us to naturally move from the discrete propagation process to the continuous case. Specifically, the embeddings $H_t$ at time $t$ incorporates all the information propagated up to time $t$ with the initial embeddings $H_0$. We can view this as a Riemann sum of an integral from time $0$ to time $t$, which leads us to the following proposition. To derive the equation using Taylor expansion, we have:
\begin{equation}
    \frac{dH(t)}{dt} = (A-I)H(t) + H(0)
\end{equation}
And,
\begin{equation}
    H(t_1) = H(0) + \int^{t_1}_{0}(A-I)H(t)dt
\end{equation}
where $H(0)$ represents the initial embeddings of all nodes, $H(t_1)$ represents the node embeddings at time $t_1$ and $A$ represents the adjacency matrix of the graph. Typically, $t_0$ is set to 0 and $t_1$ can be treated as the number of layers. Therefore, $H(t_1)$ can be interpreted as the node embeddings after $t_1$ convolutions.

\subsection{Graph Neural Ordinary Differential Equations (GODEs)}
Unlike Continuous Graph Neural Networks (CGNNs), which develop a continuous message-passing layer, Graph Neural Ordinary Differential Equations (GODEs)\cite{poli_graph_2021} parametrize the derivative function using 2 or 3 GCN layers directly. Based on graph spectral theory, the residual version of GCN layers\cite{kipf_semi-supervised_2017,Kenta_graph} are in the following form:
\begin{equation}
    H_{n+1} = \sigma(AGG(W, A, H_n))
\end{equation}
where $A$ is the graph Laplacian and $\sigma$ is a nonlinear activation function, $W$ is the transformation matrix, and $H_{n+1}$ is the node embedding at the $(n+1)$th layer. We denote the graph convolution operator as $G_C$. Where $G_C(H(n+1)) = \sigma(AGG(W, A, H_n))$. The $n$-layer convolution can be written as follows:
\begin{equation}
    H(n) = GCN(n,H(0)) = G_C^n \circ G_C^{n-1} \circ ... \circ G_C^1 H(0)
\end{equation}
Where $G_C$ is the graph convolution layer, $n$ is number of layers. And the GODEs to calculate the final node embeddings is:
\begin{equation}
    H(t_1) = H(0) + \int^{t_1}_{t_0} GCN(n, H(t)) dt
\end{equation}
where $H(t)$ represents the node embeddings at time $t$, $H(0)$ represents the initial embeddings, $GCN(n,H(t))$ represents the $n$ graph convolution layer with input node embeddings $H(t)$. $H(t_1)$ can be treat as the final node embeddings that estimate by the ODE, with the input of $H(t)$.

\subsection{ODE Solver}
There are various ODE solvers can solve the integral problem, and it can generalize various neural network architectures\cite{Chen_node}. For instance, the explicit Euler method can be written as following step:
\begin{equation}
    H(t+1) = H(t) +hf(H(t),t;\theta)
\end{equation}
 where $H(t)$ is node embeddings at state $t$. To estimate the solution of the differential equation, we can integrate the states up to $H(t_1)$ using this formula, starting from a given initial value of $H(t_0)$. The resulting $H(t)$ provide an approximation of the solution to the differential equation. Note that this equation is identical to a residual connection when $h = 1$. Other ODE solvers like RK4 it is also can be viewed as dense convolution networks\cite{Chen_node}. Other than Euler and RK4, there are many other solvers exist, such as dopri5, dopri8, fixed adams, midpoint and so on. For GODEs, higher-order ODE solvers are generally more effective, as long as the graph is dense enough to benefit from the additional computation \cite{poli_graph_2021}. In this study, we will only test two solvers: explicit Euler and RK4.

\subsection{Collaborative Filtering(CF)}
Collaborative filtering(CF)\cite{chen_revisiting_2020,chen_survey_2018,wang_neural_2019,rendle_bpr:_2009,he_neural_2017, yang_survey_2014, ebesu_collaborative_2018, wu_collaborative_2016} is a technique that can filter out items that a user might like on the basis of reactions by similar users. A typical CF paradigm involves learning user and item node embeddings, from historical interaction data, and give a top-$k$ recommendation based on the pairwise similarity of the user and item.

Let $E_0^u \in R^{N\times D}$ and $E_0^i \in R^{M\times D}$ be the initial user and item embeddings, respectively. Since user-item interactions can be represented as  bipartite graphs, it is popular to adopt GCNs for CF. One of the early used GCN-based CF methods is NGCF\cite{wang_neural_2019}, which utilizes non-linear activations and transformation matrices to map the user embedding space to the item embedding space. At each layer, user and item embeddings are combined by concatenation. However, non-linear activations and transformation matrices are not necessary for CF, as the user-item graph is sparse and non-linear GCNs are prone to overfitting. Moreover, empirical evidence suggests that transforming between user and item embedding spaces does not improve CF accuracy significantly. A simpler and more effective GCN-based CF method, called LightGCN\cite{he_lightgcn:_2020}, has been proposed. LightGCN removes the activation function and transformation matrix, and achieves state-of-the-art accuracy on many datasets. This suggests that linear GCNs with layer combination perform best among various design choices. The definition of the linear graph convolutional layer is as follows:
\begin{equation}
\begin{split}
    E_k^u &= A^kE_{k-1}^i,\\
    E_k^i &= A^kE_{k-1}^u
\end{split}
\end{equation}
Here, $A$ represents the normalized adjacency matrix, and $E_{k}$ represents the user/item embedding matrix at the $k$-th layer. The initial embeddings, which are denoted as $E^u_0$ and $E^p_0$. The final embeddings $E^u_{f}$ and $E^p_{f}$ are calculated using layer combination, expressed as follows:
\begin{equation}
\begin{split}
    E^u_{f} &= \sum^K_{l=0}w_lE_l^u, \\ 
    E^i_{f} &= \sum^K_{l=0}w_lE_l^i
\end{split}
\end{equation}
where $K$ is the number of layers, $w_l$ is weight for $l$th layer, and $E^u_{f}$ and $E^i_{f}$ are the final node embeddings. All CF methods learn the initial embeddings of users and items. The model prediction is defined as the inner product of user and item final embeddings:
\begin{equation}
    y_{u,i} = E_{f}^{u^T} E_{f}^i
\end{equation}
where $E_{f}^{u}$ is the final users embeddings and $E_{f}^i$ is the final items embeddings.
Typically, the following Bayesian personalized ranking (BPR)\cite{rendle_bpr:_2009} loss is used to train the initial embedding vectors in the field of collaborative filtering.

\subsection{Learnable-Time ODE-based Collaborative Filtering(LT-OCF)}
The aggregation of messages in LightGCN, a linear GCN model, can be viewed as the integration of an ordinary differential equation (ODE). Building on this concept, LT-OCF\cite{choi_lt-ocf:_2021} was developed as an ODE-based method. The complete model can be viewed as a continuous version of LightGCN with a learnable number of layers. Figure 1 offers a comparison of the two methods.  The message passing of LT-OCF for single ODE can be written as follows:
\begin{equation}
\begin{split}
    u(K) &= u(0) + \int^K_0 (A-I)i(t) dt, \\
    i(K) &= i(0) + \int^K_0 (A-I)u(t) dt
\end{split}
\end{equation}
where, $u(t) \in R^{N\times D}$ refers to user embeddings, and $i(t) \in R^{M\times D}$ refers to item embeddings at state $t$. $K$ can be treated as the number of layers. $u(K)$ and $i(K)$ are the user embeddings and item embeddings at time $K$, respectively. For the whole model, the time interval $[0,K]$ is split into $T$ parts. For instance, if $T=2$, the model will have $2$ ODE blocks. The first ODE block at time $[0,t_1]$ and the second ODE block at time $[t_1, K]$. The final embedding will be a combination of the output of these two blocks, the initial embeddings, and the embeddings at time $K$. The final embeddings are calculated as follows:
\begin{equation}
\begin{split}
    E^u_{f} &= w_0u(0) + \sum^T_{l=1}w_lu(t_l)+w_Ku(K),\\
    E^i_{f} &= w_0i(0) + \sum^T_{l=1}w_li(t_l)+w_Ki(K)
\end{split}
\end{equation}
where $u(0)$ and $i(0)$ are initial user embeddings and initial item embeddings respectively, $w_i$ is weight for each ODE block, and $K$ is the ending time or number of layers. $E^u_{f}$ and
$E^i_{f}$ are final user embeddings and final item embeddings, respectively.

\section{Proposed Method}
\subsection{Motivation}
As we previously mentioned, the linear GCN model can be interpreted as an ODE problem \cite{choi_lt-ocf:_2021,Chen_node}. This means that we can view the GCN process as a continuous flow of information across the graph, where the state of each node changes continuously over time, governed by a differential equation. This concept has led to the development of Graph Neural Ordinary Differential Equations (GODEs), which extend the GCN framework to a continuous-time setting.
With GODEs, we can estimate the final node embedding with one or two GCN layers. Specifically, by solving the ODEs governing the GCN process, we can obtain a continuous-time dynamical system that describes the evolution of the node embeddings over time. This can be viewed as a generalization of the static GCN model, where we can capture more complex temporal dynamics in the graph. In addition, we can also use the GODE framework to incorporate other types of information, such as node attributes or edge features. Inspired by the success of GODE in modeling continuous-time dynamics, we believe that this framework is also suitable for the collaborative filtering (CF) task. In CF, the goal is to estimate user and item embeddings. By leveraging GODEs, we can estimate these embeddings more accurately.
\subsection{Method}
We propose a new method based on the idea of Graph Neural ODE, as shown in Fig 1. Instead of creating a continuous message passing layer, we parametrize the derivative function using multiple layers of GCN directly. In other words, we use the information captured by one or two GCN layers to estimate the final state of the embedding by solving an ODE problem without creating all of the layers. Unlike the concept of CGNN\cite{deng_continuous_2019}, we treat $t_1$ as  hyperparameters, and our model doesn't have any layer combinations, since integration can be viewed as the summation of all layers from time $0$ to $t_1$. The initial embeddings are directly fed into the ODEs, and the output of the ODEs becomes the final embedding. The overall formula can be written as:
\begin{equation}
\begin{split}
    E^u_{f} = E^u(t_1) &= E^u(0) + \int^{t_1}_{0} W(A^n-I)E^i(t) dt\\
    E^i_{f} = E^i(t_1) &= E^p(0) + \int^{t_1}_{0} W(A^n-I)E^u(t) dt
\end{split}
\end{equation}
In this model, $A$ is the adjacency matrix, $E^u_0$ and $E^i_0$ are the initial user and item embeddings, respectively. $W$ is a trainable weight and $n$ is the number of layers. The user and item embeddings at time $t$ are denoted as $E^u(t)$ and $E^i(t)$, respectively. The final user embeddings and item embeddings are denoted as $E^u_{f}$ and $E^i_{f}$, respectively.
Unlike other GCN-based models \cite{he_neural_2017,he_lightgcn:_2020}, where the final embeddings are a combination of all the layers. Our method aims to estimate the final embeddings using the information that captured by several GCN layers through an ODE function. Same as other methods, we use BPR\cite{rendle_bpr:_2009} loss to train the embeddings and use ODE solvers, such as explicit Euler and RK4, to solve the ODE problem. 

When it comes to solving ODEs, there are many options available, each with their own strengths and weaknesses. In fact, different ODE solvers may produce vastly different results even when applied to the same dataset. As a result, we believe that it is crucial to carefully consider the choice of ODE solver based on the characteristics of the dataset being used. By doing so, we can ensure that we are using an appropriate solver that will provide accurate and reliable results. 

And the model prediction is defined as the inner product of the final embeddings of the user and item:
\begin{equation}
    y_{u,i} = {E^u_{f}}^TE^i_{f}
\end{equation}
where $y_{u,i}$ is the rating matrix for all users.
\subsection{Connection with CGNN and LT-OCF}
Suppose the following setting in our method:i) the number of layers $n$ is $1$, ii) We use euler's method to solve the ODE problem, iii) we set $W$ to $1$. Then our model can be written as follows:

\begin{equation}
\begin{split}
    E^u(t_1) &= E^u_0 + \int^{t_1}_{0} (A-I)E^i(t) dt \\
    E^i(t_1) &= E^i_0 + \int^{t_1}_{0} (A-I)E^u(t) dt
\end{split}
\end{equation}
which is same as the Continuous Graph Neural Networks(CGCN), or a LT-OCF with single ODE block ($K=1$).

\section{Experiment}
\begin{table*}[!ht]
    \centering
    \caption{The statistics of the datasets}
    \begin{tabular}{cc|c|c|c}
    \hline
        Datasets & \# of interactions for training & \# of interactions for validation & \# of interactions for testing & Sparsity \\ \hline
        Office & 43448 & 4905 & 4905 & 0.44867\% \\ 
        Health & 269137 & 38609 & 38609 & 0.0484\% \\ 
        Cell Phone & 138681 & 27879 & 27879 & 0.0668\% \\ 
        Beauty & 153776 & 22363 & 22363 & 0.07335\% \\ \hline
    \end{tabular}
    \label{table:data}
\end{table*}

\subsection{Experimental Environments}
In this section, we introduce our experimental environments and results. All experiments were conducted in the following hardware environments: 12th Gen Intel(R) Core(TM) i9-12900F, with one GeForce RTX 3080 GPU. We utilize the \textit{torchdiffeq} PyTorch package to solve and backpropagate through the ODE solver.

\subsection{Datasets and Baselines}
\textbf{Datasets.} We use the public Amazon Reviews dataset\cite{mcauley_image-based_2015} with four benchmark categories, including: \textit{Beauty}, \textit{Health}, \textit{Cell Phones}, \textit{Office Product}. The details of the datasets are summarized in Table~\ref{table:data}. We follow the 5-core setting as existing works on users and the same transformation\cite{he_neural_2017,he_lightgcn:_2020,fan_graph_2022, fan_graph_2023} of treating the existence of reviews as positives. We sort each user’s interactions chronologically and adopt the leave-one-out setting, with the last interacted item for testing and the second last interaction for validation.

\textbf{Eveluation.} We use the standard ranking evaluation metrics, including Recall@N and NDCG@N, to evaluate the averaged ranking performance over all users. These metrics are widely used in existing works, such as \cite{he_neural_2017, he_lightgcn:_2020, fan_graph_2022, choi_lt-ocf:_2021}. Recall@N measures the correctness of ranking the ground-truth items in the top-N list, regardless of the ranking position. It is a widely used metric for evaluating the effectiveness of top-N recommendation algorithms. NDCG@N extends the idea of Recall@N by giving higher weights to higher ranking positions. It measures the quality of a ranking algorithm based on the graded relevance of the items in the ranked list. We only report the N=20 due to limited space. 

\textbf{Baselines.} In total, we compare GODE-CF with various types of the state-of-the-art models:
\begin{itemize}
\item NGCF\cite{he_neural_2017} is a GCN-based CF method. It uses feature transformation and non-linear activations
\item LightGCN\cite{he_lightgcn:_2020} are linear GCN-based CF methods.
\item UltraGCN\cite{mao_ultragcn:_2021} is an ultra-simplified formulation of GCN which skips explicit message passing and directly approximate the limit of infinite message passing layers.
\item layerGCN\cite{zhou_layer-refined_2022} is a layer-refined GCN model, that refines layer representations during information propagation and node updating of GCN.
\item GTN\cite{fan_graph_2022} a graph trend filtering networks framework  to capture the adaptive reliability of the interactions between users and items. Extensive experiments on four real-world datasets demonstrate the.
\item LT-OCF\cite{choi_lt-ocf:_2021} is a NODE-based method that learn the optimal architecture of the model for CF.
\end{itemize}

\subsection{Hyper-Parameters Grid Search}
In this section we introduce the grid search of the hyperparameters. The learning rate is $0.001$, and the embedding size is $128$ for all models and the Normal distribution of $N(0,1)$ is used to set initial embeddings.
\begin{itemize}
    \item For LT-OCF, the regularization coefficient $\lambda$ in all method is in $\{1e-4, 1e-3, 1e-2\}$. The number of elements $K$ is in $\{2, 3, 4\}$. The number of learnable intermediate time points T is in $\{1,2,3\}$, we consider the following solvers: the explicit Euler, and RK4.
    \item  For NGCF, we search the node and message dropouts from $\{0.1, 0.3, 0.5\}$ and number of layers from $\{1,2,3,4,5\}$.
    \item For GTN, we search parameters $\beta$ and $\gamma$ from $\{0.001, 0.005, 0.01, 0.05, 0.1\}$.
    \item For UltraGCN, we search their weights from $\{1, 1e-3, 1e-5, 1e-7\}$ and negative weights from $\{10, 50, 100\}$. 
    \item For layerGCN, we search number of layers from $\{1,2,3,4,5\}$
    \item For LightGCN, we search the node and message dropouts from $\{0.1, 0.3, 0.5\}$ and number of layers from $\{1,2,3,4,5\}$
    \item For GODE-CF, we search number of layers from $\{1,2,3\}$, and $t$ from $\{0.7, 0.75, 0.8, 0.85, 0.9, 0.95, 1\}$
\end{itemize}

\subsection{Experimental Results}
In Table~\ref{Table:overall}, we summarize the overall accuracy in terms of Recall and NDCG. 
\begin{itemize}
    \item Among GCN-based methods, GODE-CF consistently achieves the best performance across all four datasets. In particular, on \textit{Health}, GODE-CF significantly outperforms the strongest GCN-based baseline, GTN, by 11\% on Recall@20. For other baselines, LightGCN exhibits the best performance on \textit{Beauty}, while GTN demonstrates the best performance on \textit{Health} compared to other GCN-based baselines.  
    \item Among all baselines, the ODE-based model LT-OCF shows state-of-the-art accuracy for \textit{Beauty}. However, our method, GODE-CF, achieves the best accuracy in all cases. The Euler solver yields the best results for all four datasets compared to RK4. We believe that the choice of solver may affect performance depending on the dataset.
\end{itemize}
We compare the training curve of LightGCN and GODE-CF for \textit{Beauty} and \textit{Health}. In general, our method provides a faster training speed in terms of the number of epochs than LightGCN.

\begin{figure}[htbp]
\centerline{\includegraphics[scale=0.57]{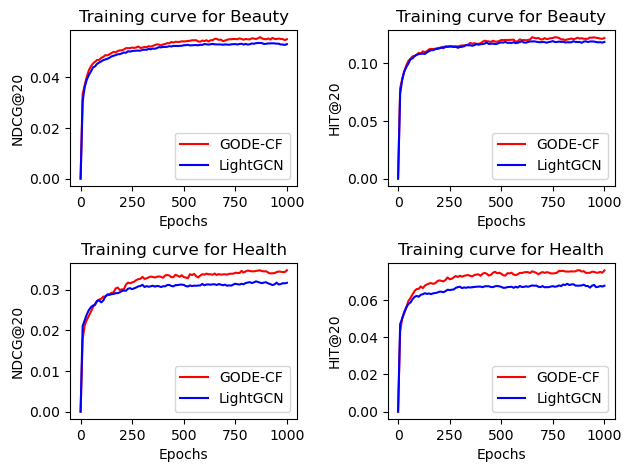}}
\caption{Training curve for Beauty and Health}
\label{fig:curve}
\end{figure}

\begin{table*}[!ht]
    \centering
    \caption{Overall Performance}
    \begin{tabular}{c|cc|cc|cc|cc}
    \hline
        Dataset &  \multicolumn{2}{c}{Beauty} &  \multicolumn{2}{c}{Health} &  \multicolumn{2}{c}{Cell Phone} &  \multicolumn{2}{c}{Office} \\ \hline
        Method & Recall@20 & NDCG@20 & Recall@20 & NDCG@20 & Recall@20 & NDCG@20 & Recall@20 & NDCG@20 \\ \hline
        NDCG & 0.070787 & 0.029954 & 0.030641 & 0.012258 & 0.043868 & 0.016914 & 0.050968 & 0.022137 \\ 
        layerGCN & 0.076197 & 0.031435 & 0.024528 & 0.010094 & 0.039671 & 0.015078 & 0.047095 & 0.021048 \\ 
        UltraGCN & 0.056611 & 0.026183 & 0.031702 & 0.013475 & 0.036049 & 0.015575 & 0.046891 & 0.023363 \\ 
        GTN & 0.071457 & 0.030585 & 0.032868 & 0.013517 & 0.045590 & 0.017802 & 0.043629 & 0.020263 \\ 
        LightGCN & 0.077762 & 0.032996 & 0.030304 & 0.012023 & 0.044299 & 0.016609 & 0.056473 & 0.026355 \\ 
        LT-OCF(Euler) & 0.078791 & 0.033095 & 0.030226 & 0.011975 & 0.046415 & 0.017393 & 0.056269 & 0.025960 \\ 
        LT-OCF(RK4) & 0.078254 & 0.033194 & 0.030330 & 0.011930 & 0.045626 & 0.017161 & 0.055250 & 0.025634 \\ \hline
        GODE-CF(RK4) & 0.078701 & 0.033480 & 0.033619 & \textbf{0.013565} & 0.048531 & 0.018325 & 0.055861 & 0.025178 \\ 
        GODE-CF(Euler) & \textbf{0.080758} & \textbf{0.034068} & \textbf{0.033878} & 0.013340 & \textbf{0.050791} & \textbf{0.019095} & \textbf{0.056677} & \textbf{0.027020} \\ \hline
    \end{tabular}
    \label{Table:overall}
\end{table*}

\subsection{Efficiency Comparison}
In this section, we provide empirical evidence demonstrating the superiority of GODE-CF's training efficiency compared to other baselines. We believe that it is essential to prove this convincingly, so we compare the total training time and epochs required to achieve the best performance among the models.
Our experiment includes validation time in the training time to provide a comprehensive overview of the training process. 
To ensure a fair comparison, we use a uniform code framework that we implemented ourselves for all models. We train all models with a fixed number of 1000 epochs to eliminate the effect of varying epoch numbers. Additionally, we conduct our experiments on all four datasets using the same machine:12th Gen Intel(R) Core(TM) i9-12900F, with one GeForce RTX 3080 GPU for all compared models. 

Table~\ref{table:efficiency} shows the details of the results. The explicit Euler providing the best performance for our model for all four datasets. Since this solver is simple and fast, which is a contributing factor to GODE-CF's short training time.
By comparing the total training time and epochs required to achieve the best performance, we demonstrate that GODE-CF consistently outperforms other baselines.

\begin{table}[!ht]
    \centering
    \caption{Efficiency comparison}
    \begin{tabular}{ccccc}
    \hline
        ~ & ~ & Training Time & ~ & ~ \\ \hline
        Dataset & Beauty & Health & Cell Phone & Office \\ \hline
        NDCG & 39m & 90m & 43m & 7m \\ 
        UltraGCN & 54m & 93m & 52m & 14m \\ 
        GTN & 73m & 207m & 67m & 8m \\ 
        LightGCN & 40m & 84m & 42m & \textbf{6m} \\ 
        LT-OCF(Euler) & 48m & 115m & 48m & 8m \\ 
        LT-OCF(RK4) & 98m & 263m & 94m & 13m \\ 
        GODE-CF(RK4) & 57m & 146m & 56m & 9m \\ 
        GODE-CF(Euler) & \textbf{35m} & \textbf{80m} & \textbf{36m} & 7m \\ \hline
    \end{tabular}
    \label{table:efficiency}
\end{table}

\section{Ablation Studies}
\subsection{Euler vs. RK4}
\begin{figure}[htbp]
\centerline{\includegraphics[scale=0.57]{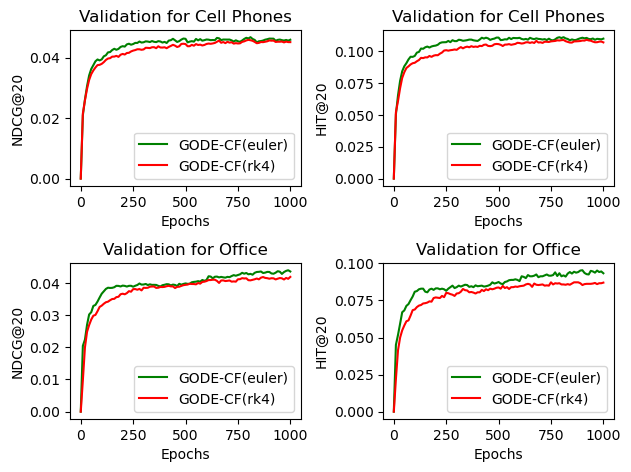}}
\caption{Training curve for \textit{Cell phones} and \textit{Office} with different solvers}
\label{fig:solver1}
\end{figure}

\begin{figure}[htbp]
\centerline{\includegraphics[scale=0.57]{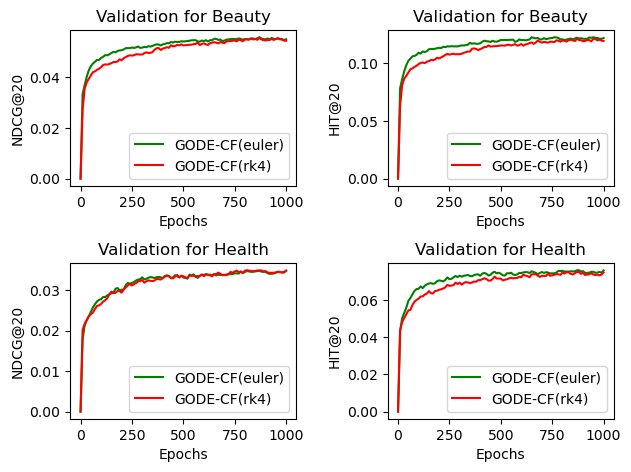}}
\caption{Training curve for \textit{Beauty} and \textit{Health} with different solvers}
\label{fig:solver2}
\end{figure}

In this section, we compare two different solvers for ordinary differential equations (ODEs): Euler and RK4. The Fig ~\ref{fig:solver1} and Fig~\ref{fig:solver2} are the training curves of these two solvers. Other solvers, such as DOPRI, are almost the same as RK4, but RK4 has smaller computation complexity. Therefore, we only use RK4 and Euler’s method for this experiment. Contrary to what we thought, Euler's method shows better accuracy for almost all cases, whereas RK4 should have better accuracy in terms of solving general ODE problems. For instance, the Euler method achieves a Recall/NDCG of 0.050791/0.019095 vs. 0.048531/0.018325  with RK4 on \textit{Cell Phones}. However, for other datasets, we observe very similar patterns where Euler's method consistently outperforms RK4. And Euler's method provides a faster training speed in terms of the number of epochs than that of RK4 for all of the cases. This is because using the RK4 solver is more likely to result in overfitting the training data in this model.

\subsection{Impact on different number of layers}
We also conducted experiments that explored the impact of varying the number of layers inside ODE on the model's performance. The detailed results are in Table~\ref{table:layers}. 
\begin{table*}[!ht]
    \centering
    \caption{Performance comparison for different numbers of layers}
    \begin{tabular}{ccc|cc|cc|cc}
    \hline
        Dataset &  \multicolumn{2}{c}{Beauty} &  \multicolumn{2}{c}{Health} &  \multicolumn{2}{c}{Cell Phone} &  \multicolumn{2}{c}{Office} \\ \hline
        ~ & Recall@20 & NDCG@20 & Recall@20 & NDCG@20 & Recall@20 & NDCG@20 & Recall@20 & NDCG@20 \\ \hline
        1 layer & 0.07526 & 0.03093 & 0.02901 & 0.01104 & 0.04204 & 0.01568 & 0.05647 & 0.02522 \\ 
        2 layer & \textbf{0.08076} & \textbf{0.03407} & \textbf{0.03388} & \textbf{0.01356} & \textbf{0.05079} & \textbf{0.01909} & \textbf{0.05668} & \textbf{0.02702} \\ 
        3 layer & 0.07919 & 0.03311 & 0.03124 & 0.01217 & 0.04760 & 0.01783 & 0.05770 & 0.02639 \\ \hline
    \end{tabular}
    \label{table:layers}
\end{table*}

\begin{table*}[!ht]
    \centering
    \caption{Performance with or without weight}

    \begin{tabular}{lll|ll|ll|ll}
 
    \hline
        Dataset &  \multicolumn{2}{c}{Beauty} &  \multicolumn{2}{c}{Health} &  \multicolumn{2}{c}{Cell Phone} &  \multicolumn{2}{c}{Office} \\ \hline
        ~ & Recall@20 & NDCG@20 & Recall@20 & NDCG@20 & Recall@20 & NDCG@20 & Recall@20 & NDCG@20 \\ \hline
        Without weight (W) & \textbf{0.08076} & \textbf{0.03407} & \textbf{0.03403} & 0.01353 & 0.05058 & \textbf{0.01916} & 0.05586 & 0.02661 \\ 
        With weight (W) & 0.08027 & 0.03395 & 0.03362 & \textbf{0.01356} & \textbf{0.05079} & 0.01909 & \textbf{0.05668} & \textbf{0.02702} \\ \hline
    \end{tabular}
    \label{table:weight}
\end{table*}

Based on the results, we have found that a 2-layer design yields optimal results for all four cases. This is because the ODE needs to estimate the final state of the embeddings by using enough information captured by the GCN layer. In other words, the GCN layer need to provides the necessary information for the ODE to make optimal estimations.

On the other hand, a 1-layer design may not provide enough information for the ODE to make a good estimation. This is because one GCN layer may not capture all the necessary information required to estimate the final state of the embeddings.

Conversely, a 3-layer design may capture excessive information, leading to overfitting of the training data. To address this issue, we suggest incorporating the dropout technique when using a 3-layer design. Dropout can help prevent overfitting and ensure that the final state estimated by the ODE is more reliable.

\subsection{Sensitivity on t}
\begin{figure}[htbp]
\centerline{\includegraphics[scale=0.57]{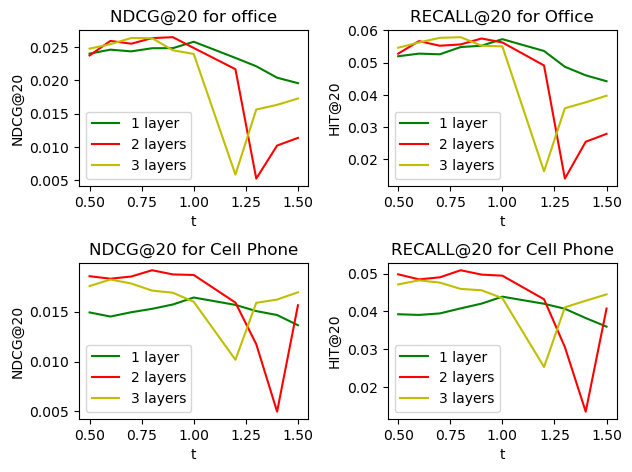}}
\caption{Performance for different $t$ on \textit{office} and \textit{Cell Phone} }
\label{fig:t}
\end{figure}

By varying $t$, we also investigate how different $t$ affect the model accuracy. The detailed results are shown in Figure ~\ref{fig:t}.
According to the plot, we can see that the model with only a small number of layers requires a larger value of $t$ to reach the optimal performance. Conversely, models with larger numbers of layers require a smaller value of $t$ to achieve the optimal performance. This is because models with fewer layers provide less information to the ODE, which results in the requirement of a longer time step to reach the final state. Conversely, models with a higher number of layers capture more information, which subsequently results in them being closer to the final state, thereby allowing the ODE to take smaller time steps $t$. It is important to note that the number of layers plays a crucial role in the performance of the model. An interesting finding is that, for both 2-layer and 3-layer settings, there is a time point at which the model experiences a significant performance drop. However, after this time point, the performance begins to recover. The cause of this issue remains unknown, so we will leave it for future work.

\subsection{Impact on weight}
We are model has a learneable weight for each layer inside ODE. We are also interested in  to investigate whether the weight has a significant impact on the model's performance. The detailed results are shown in Table \ref{table:weight}. Based on the results, we found that the impact of weight is highly dependent on the dataset. For most cases, the model without weight has better performance than the model with weight on the \textit{Beauty}, but the difference is not significant. However, for the \textit{Office}, the model has a larger improvement when using weight. Whether or not to use weight depends highly on the specific case. It's recommended to evaluate the impact of weight on the model for each specific dataset.

\section{Conclusion}
This study aims to introduce a new method for Collaborative Filtering (CF) called GODE-CF. The method is based on Graph Neural Ordinary Differential Equation (ODE) and is formulated as a graph neural ODE system suitable for collaborative filtering. We conducted an evaluation of GODE-CF on four different real-world datasets. The experimental results showed that GODE-CF outperforms various state-of-the-art baselines in terms of performance with shorter training time.
Looking into the future, we plan to test more ODE solvers as we only tested two in this study, and there are a lot more ODE solvers is being developed. Additionally, we discovered that our model is prone to overfitting, which is a concern for future research.  In conclusion, GODE-CF provides a new approach to CF, which has the potential to improve the performance and efficiency of collaborative filtering.

\section*{Acknowledgment}

This work is supported in part by NSF under grant  III-2106758. 

\bibliographystyle{IEEEtran}
\bibliography{IEEEabrv, GODE-CF}

\vspace{12pt}

\end{document}